\pdfoutput=1

\documentclass[11pt]{article}

\usepackage[final]{acl}

\usepackage{times}
\usepackage{latexsym}

\usepackage[T1]{fontenc}

\usepackage[utf8]{inputenc}

\usepackage{microtype}

\usepackage{inconsolata}
\usepackage{makecell}

\title{DART: Leveraging Multi-Agent Disagreement for Tool Recruitment in Multimodal Reasoning}

\author{
    \textbf{Nithin Sivakumaran}\textsuperscript{1} \quad\quad
    \textbf{Justin Chih-Yao Chen}\textsuperscript{1} \quad\quad
    \textbf{David Wan}\textsuperscript{1} \quad\quad
    \textbf{Yue Zhang}\textsuperscript{1} \\
    \textbf{Jaehong Yoon}\textsuperscript{2} \quad\quad
    \textbf{Elias Stengel-Eskin}\textsuperscript{3} \quad\quad
    \textbf{Mohit Bansal}\textsuperscript{1} \\
    \textsuperscript{1}UNC Chapel Hill \quad \textsuperscript{2}Nanyang Technological University\\ \textsuperscript{3}The University of Texas at Austin
}

\usepackage{amsmath,amsfonts,bm}

\def\eqref#1{equation~\ref{#1}}

\def\1{\bm{1}}

\DeclareMathAlphabet{\mathsfit}{\encodingdefault}{\sfdefault}{m}{sl}
\SetMathAlphabet{\mathsfit}{bold}{\encodingdefault}{\sfdefault}{bx}{n}

\usepackage[capitalize]{cleveref}
\AddToHook{cmd/appendix/before}{\crefalias{section}{appendix}}
\usepackage{url}
\usepackage{multirow}
\usepackage{booktabs}
\usepackage{wrapfig}
\usepackage[table]{xcolor}
\usepackage{booktabs}      
\usepackage{colortbl}      
\usepackage{adjustbox}     
\usepackage{pifont}
\usepackage{graphicx}
\graphicspath{ {./figures/} }
\usepackage{tcolorbox}

\newcommand{\method}[1]{DART}
\newcommand{\fullform}[0]{\textbf{D}isagreement among \textbf{A}gents for \textbf{R}ecruitment of \textbf{T}ools}
\newcommand{\ssymbol}[1]{^{\@fnsymbol{#1}}}

\begin{document}
\maketitle

\begin{abstract}
Specialized visual tools can augment large language models or vision language models with expert knowledge (e.g., grounding, spatial reasoning, medical knowledge, etc.), but knowing which tools to call (and when to call them) can be challenging.
We introduce \method{}, a multi-agent framework that uses disagreements between multiple debating visual agents to identify useful visual tools (e.g., object detection, OCR, spatial reasoning, etc.) that can resolve inter-agent disagreement. 
These tools allow for fruitful multi-agent discussion by introducing new information, and by providing tool-aligned agreement scores that highlight agents in agreement with expert tools, thereby facilitating discussion. We utilize an aggregator agent to select the best answer by providing the agent outputs and tool information.
We test \method{} on four diverse benchmarks and show that our approach improves over multi-agent debate as well as over single agent tool-calling frameworks, beating the next-strongest baseline (multi-agent debate with a judge model) by 3.4\% and 2.4\% on A-OKVQA and MMMU respectively. 
We also find that \method{} adapts well to new tools in applied domains, with a 1.3\% improvement on the M3D medical dataset over other strong tool-calling, single agent, and multi-agent baselines. Additionally, we measure text overlap across rounds to highlight the rich discussion in \method{} compared to existing multi-agent methods. Finally, we study the  tool call distribution, finding that diverse tools are reliably used to help resolve disagreement.\footnote{Code: \url{https://github.com/nsivaku/dart}}
\end{abstract}

\section{Introduction}
\label{sec:intro_clean}

\begin{figure*}[t]
\begin{center}
\includegraphics[width=1.0\textwidth]{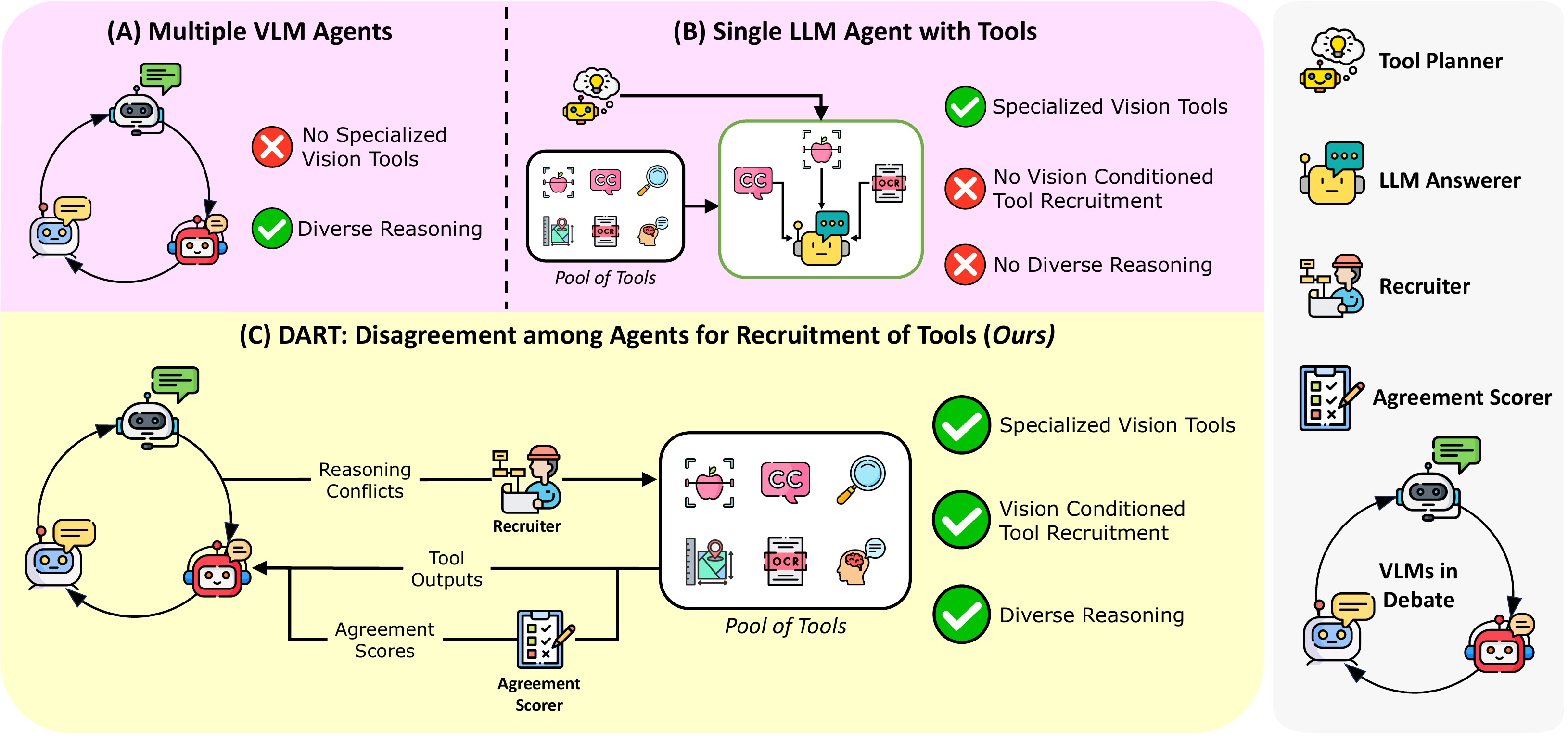}
\end{center}
\vspace{-1em}
\caption{
Previous work has explored using (A) multiple agents in  debate to refine their reasoning, but this approach is limited to the abilities of the agents. Alternatively, some methods employ a (B) top‑down LLM agent that invokes vision tools, yet they plan tool usage based solely on the question and overlook the visual information itself.
In our method (C), we facilitate a discussion among multiple agents with targeted intervention from a pool of vision tools. These tools address disagreements detected in a debate of VLM agents, with their specialized vision outputs and agreement scores being used for future discussion.}

\label{fig:1}
\end{figure*}

A key strength of human intelligence is the ability to debate and discuss our reasoning with others.
However, everyone has had the experience of engaging in a debate that goes nowhere because of a lack of new information: without the incorporation of new knowledge, each party becomes entrenched in their point of view. 
While past work \citep{wang2024rethinking, chen2024reconcileroundtableconferenceimproves, estornell2025acccollab, estornell2024multillm} has found that large-language model agents can improve their performance by engaging in multi-agent debate, this has largely been explored in unimodal, text-only settings. 
We find that when extending debate to multimodal settings with vision-language models (VLMs), the discussion between agents frequently becomes unproductive.
The lack of progress in conversation is in part due to a lack of new information, with the VLM-based agents participating in the discussion often being unable to resolve perceptual differences (see \cref{fig:1}.A).
Moreover, each VLM suffers from similar perceptual limitations: past work has found that VLMs often lag behind specially-trained expert models in domains like grounding, OCR, and spatial reasoning \citep{chiu2025aideagenticallyimprovevisual, spatialvlm}. 
Indeed, because of these limitations, a separate line of work has explored using LLMs to call expert models as tools rather than relying on VLMs (see \cref{fig:1}.B) \citep{yang2023mmreact, lu2023chameleon, wu2023visualchatgpttalkingdrawing, surismenon2023vipergpt, mialon2023augmentedlanguagemodelssurvey,yu2025mexa}, benefiting from the reasoning abilities of LLMs and the visual perceptual abilities of expert models.

We unify these three approaches (VLMs, multi-agent debate, and tool calling) with \fullform{} (\method{}), a multi-agent approach which approaches visual question-answering (VQA) tasks by engaging VLMs in a multi-agent debate, resolving the disagreement between agents with information from expert tools (see \cref{fig:1}.C). 
This has several key advantages: first, the novel information from expert tools helps move the discussion between agents forward productively and augments their perceptual abilities.
Second, it simplifies the tool-calling process; unlike top-down LLM-based solutions which plan what tools to execute in a single step based only on the question -- e.g., ViperGPT \citep{surismenon2023vipergpt}, Chameleon \citep{lu2023chameleon} -- 
\method{} adaptively calls tools based only on the points which the VLMs disagree on.
This has the added benefit of combining the strong reasoning abilities of VLMs (backed by LLMs) with the specialized perception of vision tools. 
Finally, the added information given by tools serves as an intuitive measure of confidence in each agent: the more a given agent agrees with the expert tools (which have stronger perceptual abilities), the more it can be relied on. 

\begin{figure*}[t]
\begin{center}

\includegraphics[width=1.0\textwidth]{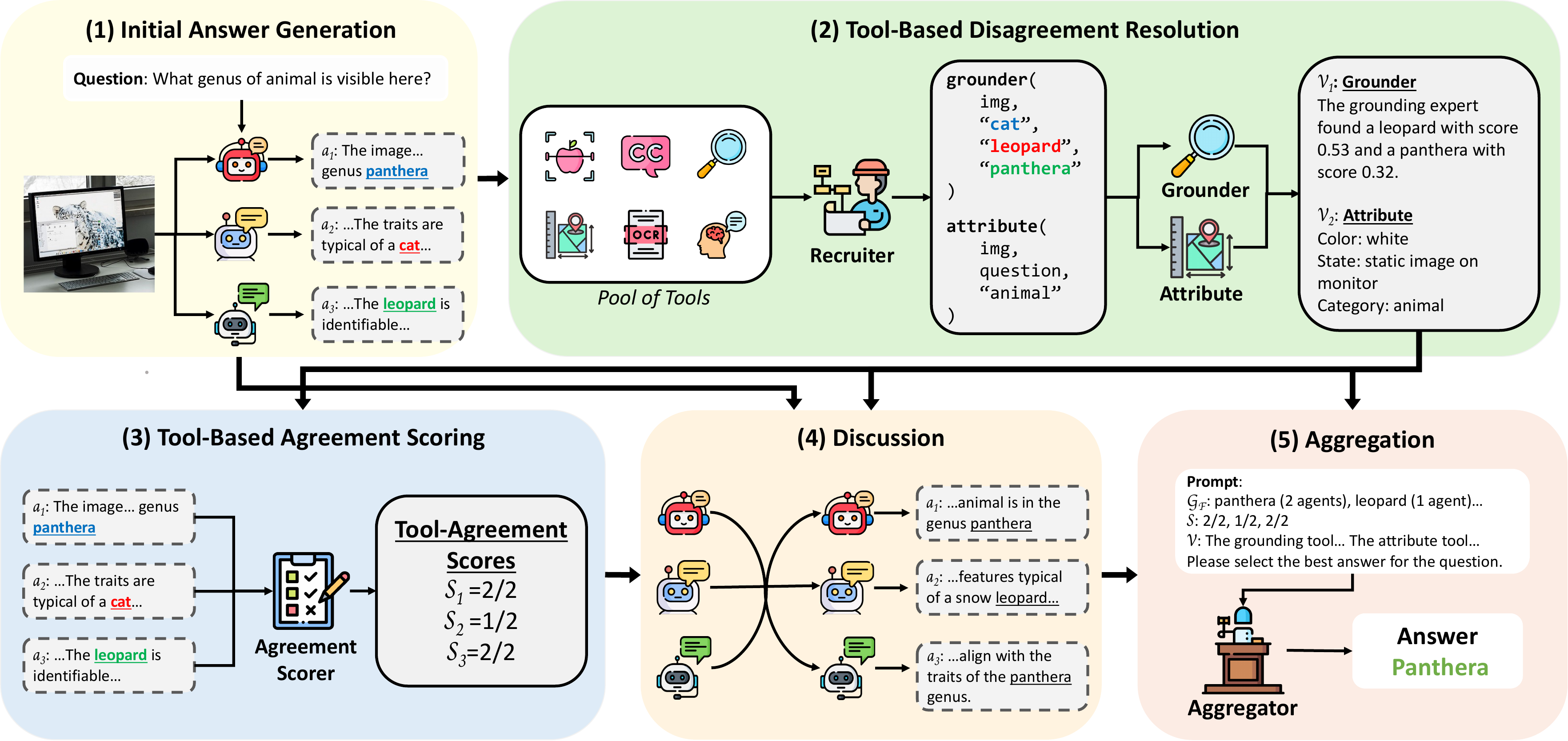}
\end{center}
\vspace{-1em}
\caption{Overview of \method{}.  We start with (1) \emph{Initial Answer Generation} from a set of answering/reasoning agents. This is followed by (2) \emph{Tool-Based Disagreement Resolution} and (3) \emph{Agreement Scoring}. The newly generated tool outputs and agreement scores are incorporated into the (4) \emph{Discussion} and (5) \emph{Aggregation} phases.}
\label{fig2:pipeline}
\end{figure*}

\method{} consists of five steps, outlined in \cref{fig2:pipeline}. \method{} starts with (1) \emph{Initial Answer Generation}, where each agent creates an answer and reasoning for an image-question pair. Next, our method uses the answers and reasoning chains from the answering agents in (2) \emph{Tool-Based Disagreement Resolution}, which aims to resolve disagreements that occurred during the initial answering step by invoking tools based on the identified conflicts. Then, in (3) \emph{Tool-Based Agreement Scoring}, \method{} compares the expert tool outputs and agent answers to calculate agreement scores. These new tool outputs and per agent agreement scores are used during (4) \emph{Discussion} where agents refine their own answer and reasoning. 
Finally, \method{} composes the tool outputs, recalculated agreement scores, and post-discussion answer outputs into a final answer in the (5) \emph{Aggregation} step.

We test \method{} on four different benchmarks that require strong and diverse perception capabilities and reasoning. We find that \method{} improves over the next best baseline (multi-agent debate with a judge model) by $3.4\%$ on A-OKVQA \citep{schwenk2022aokvqa}, $2.4\%$ on MMMU \citep{mmmu}, and $1.2\%$ on NaturalBench \citep{naturalbench}. 
Similarly, DART outperforms tool-calling baselines like ViperGPT \citep{surismenon2023vipergpt}
and Chameleon \citep{lu2023chameleon} on A-OKVQA, MMMU, and NaturalBench by $19.0\%$, $11.5\%$, and $5.3\%$ respectively.
We also test how \method{} adapts to a subject-specific domain,
adding in a relevant domain expert tool. Specifically, we test \method{}'s adaptability to medical data by testing on the M3D dataset \citep{bai2024m3d} while adding in MedGemma 4B \citep{medgemma-hf} into the expert tool set. We find that our system improves performance on M3D by $1.3\%$ over Qwen2.5-VL \citep{Qwen2.5-VL}, the strongest single agent baseline, and achieve positive gains in a multi-agent setting compared to naive multi-agent debate (which decreases by 2.7\% compared to Qwen2.5-VL). 
Our results indicate that \method{} leads to stronger discussion, introducing more novel ideas into the discussion round compared to multi-agent debate. Moreover, \method{} benefits from calling a variety of tools based on reasoning disagreements from multiple models, without overly relying on any singular tool. 
\section{Method}
\label{sec:method}

The problem setup is a Visual Question Answering (VQA) task where we use a set of VLM agents and tools in \method{} to arrive at a final answer.
The framework takes in an image and question as input and generates a single natural language answer.
Formally, given a question $Q$ and image $I$, we aim to generate a single answer $a$ based on a discussion among generalist answering agents $\textbf{A} = \{A_i\}_{i=1}^n$ and vision expert tools $\textbf{V}=\{V_i\}_{i=1}^m$.

\subsection{Initial Answer Generation}
\label{subsec:initial}
In initial answer generation (see \cref{fig2:pipeline} (Step 1), we prompt each answering agent $A_i$ to use CoT to generate a candidate answer $a_i$, a corresponding reasoning chain $r_i$, and self-reported confidence score $c_i$.
Before being passed to future stages of the pipeline, we group answers \(\{a_i\}_{i=1}^n\), reasoning \(\{r_i\}_{i=1}^n\),  and confidence \(\{c_i\}_{i=1}^n\) following \citet{chen2024reconcileroundtableconferenceimproves} to form $G_I$. The grouping process lists each unique agent answer with the number of agents supporting it and their reasoning.

\subsection{Tool-Based Disagreement Resolution}
\label{subsec:disagree}
We expect answering agents with distinct outputs to have reasoning chains that disagree at specific points. These disagreements mean that agents diverge in their perception or analysis of the scene and question.
As such, we aim to address these divergences with tool-based disagreement resolution to directly provide visual information from relevant vision expert tools that will ground future conversations  (see \cref{fig2:pipeline} (Step 2). We detect disagreements among the agents' answers \(\{a_i\}_{i=1}^n\) and answer reasoning chains \(\{r_i\}_{i=1}^n\) using a recruitment agent.
 For each of our vision tools, the recruitment agent is parameterized by an LLM provided with the acceptable input parameters (either none, text query, list of text queries) as well as potential use cases for each tool in the prompt (e.g., \emph{the grounder can be used if an agent mentioned an object that another agent missed}).
Then, given a grouped answer $G_I$, the recruitment agent detects disagreements and proceeds to construct a vision tool set and corresponding input parameters. The recruitment agent is instructed to state the disagreement and justify how a given tool would address that disagreement (see \cref{fig:disagreement} for the prompt).
For example, if the recruitment agent finds that ``\textit{Agents disagree on the color and specific markings of the planes, with one agent mentioning a green color while others describe the distinctive blue,}'' the recruitment agent would call on the attribute detection tool to ``\textit{clarify the colors and markings of the planes.}'' \method{} uses grounding, object detection, OCR, spatial reasoning, captioning, and attribute detection tools.

The grounding, object detection, OCR, and spatial reasoning tools are implemented as expert, finetuned models.
In contrast, reasoning, captioning, and attribute detection are handled by a general-purpose VLM instructed to focus on those specific tasks without leveraging task-specific models. 
We use a separate VLM from the debate agents to avoid bias, as using the same VLM might bias the output towards that particular VLM's idiosyncrasies.
 
The vision tools process image $I$ with their respective generated input parameters to get expert tool outputs $\mathcal{E} = \{E_i\}_{i=1}^l$. Note that certain vision tools, such as grounding, object detection, and OCR, are not naturally suited for natural language outputs.
For tools that do not natively output natural language (Grounder, Object
Detector) we perform minimal post-processing to convert outputs into statements suitable for language-model consumption (see \cref{fig:disagreement} for the prompt).

\subsection{Tool-Based Agreement Scoring}
\label{subsec:conf}
Domain-specific models have been shown to be stronger than VLMs within their specific skill set \citep{chiu2025aideagenticallyimprovevisual, spatialvlm}.
As such, we hypothesize that VLMs whose outputs agree more closely with those of specialized tool models are more likely to be accurate. The expert tools have better perception, meaning VLMs that match the experts are likely to have correctly perceived the image.
Thus, these agreement scores serve as an indicator for whether or not to trust a VLM output.

For every pair $((a_i,r_i),E_j)$ of output from answering agent $i$ and output from tool $j$, an agreement scorer LLM produces a
binary agreement score between answering agent $i$ and output from invoked tool $j$ as $s_{ij}\in\{0,1\}$ after comparing $(a_i,r_i)$ with the $E_j\in\mathcal{E}$ (see \cref{fig2:pipeline} (Step 3)). The agreement scorer agent produces scores that represent the degree to which an agent agrees with one specific expert tool output $E_j$.
Agreement of answering agent $i$ with the full set of tools is then represented by the mean: 
\[S_i \;=\;
\frac{1}{|\mathcal{E}|}\sum_{E_j\in\mathcal{E}} s_{ij},
\qquad
s_i\in[0,1].\]
Each $(a_i,r_i)$ is now annotated with its agreement score $S_i$ and carried into the subsequent discussion and aggregation steps. These steps provide agreement scores such that more trust is placed on agents that align with expert information. The agents that do have high scores are more likely to have accurate visual understanding of the scene. We provide  the agreement scorer prompt in \cref{fig:agreement}.

\subsection{Discussion}
\label{subsec:disc}
In the discussion, each agent takes in the grouped input $G_I$, tool outputs $\mathcal{E}$, 
and the expert-based agreement scores $S_I$  (see \cref{fig2:pipeline} (Step 4)). Recall that the grouped agent solution form includes self-reported confidence. This means that agents have both self-reported confidence and agreement scores during the discussion step. Agents are instructed to place high priority on the expert outputs, even over the agent solutions.  Answering agents are also given a 3-shot prompt with simple synthetic examples to guide them to follow the provided output format (see \cref{fig:discussion} for the full prompt). 
If the agent solutions contain information that contradicts the tool outputs, models are told to assume that the vision tool is correct. 
Finally, agents are provided with the tool-based agreement scores as well as the self-reported confidence score. 
This outputs the final group of agent answers defined as $G_F$ (see \cref{fig2:pipeline} (Step 3)), and we recalculate final confidence scores following \Cref{subsec:conf} to obtain $S_F$.

\subsection{Aggregation}
\label{subsec:aggregation}
Similar to the discussion step, we provide an aggregator agent with grouped solutions $G_F$, tool outputs $\mathcal{E}$, and tool-based agreement scores $S_F$ (see \cref{fig2:pipeline} (Step 5)). The aggregator VLM is instructed to choose, with CoT, the best answer out of all the provided solutions (see \cref{fig:aggregator} for prompt).

\section{Results and Analysis}
\label{sec:results}

\subsection{Experimental Setup}

\paragraph{Datasets and Metrics.} 
We evaluate our framework on four Visual Question Answering (VQA) datasets: \textbf{A-OKVQA} \citep{schwenk2022aokvqa}, \textbf{MMMU} \citep{mmmu}, \textbf{NaturalBench} \citep{naturalbench}, and \textbf{M3D-VQA} \citep{bai2024m3d}. We aim to test whether \method{} performs well on a dataset that focuses on reasoning and world knowledge, such as A-OKVQA. We evaluate on the direct answer subset using the metric from \citet{vqa} following the A-OKVQA direct answer evaluation scheme. 
We also test on MMMU to assess \method{}'s effect on a large, diverse set of VQA data, and on NaturalBench, which is designed to rigorously test VLMs on their visual comprehension using adversarial VQA pairs that are easy for humans but challenging for VLMs. 
Additionally, we evaluate how well \method{} adapts to a new tool by testing it on M3D-VQA, which is a medical VQA dataset on 3D medical scans of real patients. 
MMMU, NaturalBench, and M3D are all multiple choice datasets, so we measure accuracy.

\begin{table*}[t]
\centering
\small
\resizebox{0.8\textwidth}{!}{
\begin{tabular}{llccc}
\midrule
\textbf{Method} & \textbf{Agent} & \textbf{A-OKVQA} & \textbf{MMMU} & \textbf{NaturalBench} \\
\midrule
\rowcolor{gray!20}\multicolumn{5}{c}{\textit{Tool Systems}} \\
\addlinespace
ViperGPT \citep{surismenon2023vipergpt}               & Qwen               & $49.9$ & $54.0$ & $75.3$ \\
Chameleon \citep{lu2023chameleon}                & Qwen               & $47.5$ & $51.6$ & $77.2$ \\
\midrule

\rowcolor{gray!20}\multicolumn{5}{c}{\textit{Single Agent}} \\
\addlinespace
\multirow{4}{*}{CoT \citep{wei2022chain}}                           & QwenVL             & $61.3$ & $58.8$ & $79.2$ \\
                     & MiniCPM-o          & $55.7$ & $50.4$ & 77.9 \\
                         & Ovis2              & $61.0$ & $50.7$ & $78.7$ \\
                         & LLaVA 1.6 & 62.8 & 39.4 & 70.9 \\
\midrule
\multirow{4}{*}{\makecell[l]{Self-Refinement\\\citep{selfrefine}}}                      & QwenVL             & $61.8$ & $56.5$ & $79.0$ \\
          & MiniCPM-o          & $56.7$ & $46.7$ & 78.2 \\
                         & Ovis2              & $61.7$ & $47.2$ & $78.4$ \\
                         & LLaVA 1.6 & 62.9 & 39.7 & 70.6 \\
\midrule
\multirow{4}{*}{\makecell[l]{Self-Consistency (5-way)\\\citep{wang2023selfconsistency}}}                       & QwenVL             & $63.7$ & $60.8$ & $80.1$ \\
 & MiniCPM-o          & $58.2$ & $52.5$ & $78.1$ \\
                         & Ovis2              & 64.6 & 52.8 & 79.3 \\
                         & LLaVA 1.6 & 64.2 & 41.3 & 72.4 \\
\midrule
\rowcolor{gray!20}\multicolumn{5}{c}{\textit{Multiple Agents}} \\
\addlinespace
\multirow{4}{*}{Debate with Consensus}    & 3$\times$QwenVL    & $61.9$ & $60.5$ & $80.0$ \\
                         & 3$\times$MiniCPM-o & $57.4$ & $53.2$ & $78.3$ \\
                         & 3$\times$Ovis2     & $62.8$ & $57.0$ & $80.1$ \\
                         & Q, M, O            & $64.2$ & \underline{$63.1$} & $80.6$ \\
\midrule
\multirow{4}{*}{Debate with Judge}         & 3$\times$QwenVL    & $62.1$ & $60.3$ & $79.9$ \\
                         & 3$\times$MiniCPM-o & $59.7$ & $54.8$ & $78.3$ \\
                         & 3$\times$Ovis2     & $63.0$ & $58.3$ & $80.4$ \\
                         & Q, M, O            & \underline{$65.5$} & \underline{$63.1$} & \underline{$81.3$} \\
\midrule
\multirow{2}{*}{\method{} (\textit{Ours})}        & Best Model         & $65.3$ & $61.6$ & $80.8$ \\
                         & Q, M, O          & $\mathbf{68.9}$ & $\mathbf{65.5}$ & $\mathbf{82.5}$ \\
\hline
\end{tabular}
}
\vspace{-0.5em}
\caption{Comparison of \method{} and different baselines on A-OKVQA, MMMU, and NaturalBench. Q, M, and O stand for QwenVL, MiniCPM-o, and Ovis2 respectively. For the \method{} (Best Model) setting, we take the single best performing agent from multi-agent debate with judge and run \method{} with three instances of that model. The best models are Ovis2, QwenVL, and Ovis2 for A-OKVQA, MMMU, and NaturalBench respectively. 
}
\label{tab:main}
\end{table*}

\paragraph{Implementation Details.} 
We use a diverse set of answering agents and visual tools. For answering agents, we test with Qwen2.5-VL \citep{Qwen2.5-VL}, MiniCPM-o 2.6 \citep{yao2024minicpm}, and Ovis2 \citep{lu2024ovis} as our VLMs. As an additional independent single agent baseline, we include LLaVA-1.6 (Mistral) \citep{liu2024llavanext}. Then, we construct a pool of visual expert tools with GroundingDINO \citep{liu2023grounding} (grounding), YOLOv11 \citep{yolo11} (object detection), SpaceLLaVA \citep{spatialvlm} (spatial), OCR-Qwen (OCR), and InternVL-2.5 MPO \citep{wang2024mpo} (captioning, attribute, and reasoning) as vision tools. We use InternVL-2.5 to serve as an expert tool in multiple domains, as prior work has shown that models can be designed to be more effective in specific domains when prompted to do so \citep{xu2025expertpromptinginstructinglargelanguage, wang2024unleashingemergentcognitivesynergy}.  
We introduce a separate VLM for this task to reduce overlap between answering agent information and tool calling information. We also test the effectiveness of \method{} by ablating the VLM-based tools (see \cref{app:vlm_effects}).
We use Qwen2.5 (7B) as the recruitment agent and agreement scorer. 
We also find that Ovis2 serves as the best aggregator among a pool of similar VLMs. When testing on M3D-VQA, we additionally insert MedGemma 4B \citep{medgemma-hf} into the pool of  tools to serve as a medical expert.

\paragraph{Baselines.} We compare \method{} to multiple single-agent, multi-agent, and tool-calling baselines. We include comparisons to training-free single-agent tool-calling baselines: ViperGPT \citep{surismenon2023vipergpt} and Chameleon \citep{lu2023chameleon}. ViperGPT utilizes a code generation model to predict executable code that leverages required tools for a given question. In contrast to the flexibility offered by free-form code generation in ViperGPT, Chameleon adopts a more robust compositional approach, utilizing an LLM-based planner to directly invoke various tools in sequence via word matching. We update these methods to use our pool of experts and also incorporate more recent models as planners to ensure a fair comparison. 

Our single agent baselines are zero-shot CoT \citep{wei2022chain}, self-refinement \citep{selfrefine}, and self-consistency \citep{wang2023selfconsistency}. 
For multi-agent baselines, we use Debate with Consensus and Debate with Judge. 
Specifically, we follow the ReConcile LLM debate framework outlined by \citet{chen2024reconcileroundtableconferenceimproves} for both consensus and InternVL2.5 as a judge. Debate with consensus counts the most frequent answer from the agents while debate with judge uses a separate VLM model to choose an answer. See \cref{app:baselines} for more details about the baselines.

\subsection{Results} 

\begin{table*}[t]

  \centering
  \resizebox{0.9\textwidth}{!}{
  \begin{tabular}{@{} l c c c c c c @{}}
    \toprule
    \textbf{Method}     & \textbf{Plane} & \textbf{Phase} & \textbf{Organ} & \textbf{Abnormality} & \textbf{Location} & \textbf{Avg.}\\
    \midrule
    CREMA \citep{yu2025crema}       & $14.9$ & $26.7$ & $15.9$ & $17.3$ & $13.0$ & $17.2$ \\
    GPT-4o \citep{openai2024gpt4ocard}      & $83.3$ & $\underline{42.7}$ & $50.0$ & $41.3$ & $41.3$ & $51.7$ \\
    MEXA \citep{yu2025mexa}        & $65.0$ & $\mathbf{48.1}$ & $\mathbf{60.9}$ & $44.8$ & $48.0$ & $53.3$ \\
    \midrule
    Ovis2       & $59.6$ & $34.0$ & $39.8$ & $55.4$ & $43.0$ & $46.4$ \\
    MiniCPM-o   & $79.8$ & $28.4$ & $\underline{53.9}$ & $49.2$ & $37.8$ & $49.8$ \\
    QwenVL      & $\underline{91.2}$ & $33.5$ & $49.7$ & $60.0$ & $47.7$ & $\underline{56.4}$ \\
    Debate with Consensus & $81.2$ & $32.0$ & $43.1$ & $\underline{60.9}$ & $\mathbf{51.9}$ & $53.7$ \\
    Debate with Judge     & $79.5$ & $32.0$ & $45.6$ & $\mathbf{61.2}$ & $51.7$ & $53.4$ \\
    \midrule
    DART (\textit{Ours)}        & $\mathbf{92.8}$ & $40.7$ & $47.6$ & $55.9$ & $\underline{51.8}$ & $\mathbf{57.8}$ \\
    \bottomrule
  \end{tabular}
  }
  \vspace{-0.5em}
  \caption{Accuracy on the M3D medical dataset by question type and overall. \method{} performs the best out of the tested methods.}
  \label{tab:m3d}
\end{table*}

\begin{table}[h!]
\centering
\resizebox{\columnwidth}{!}{
\begin{tabular}{lcccc}
\toprule
\textbf{Method} & \textbf{R-1} & \textbf{R-2} & \textbf{R-L} & \textbf{Jac. Index} \\
\midrule
Debate with Consensus & $0.679$ & $0.529$ & $0.550$ & $0.559$ \\
\method{} & $0.452$ & $0.310$ & $0.377$ & $0.270$ \\
\bottomrule
\end{tabular}
}
\caption{Comparison of \method{} and multi-agent debate across text similarity metrics. R and Jac. Index stand for ROUGE and Jaccard Index respectively.}
\label{tab:text_overlap}
\end{table}

\paragraph{\method{} is stronger than all tool-calling, single-agent, and multi-agent baselines.}

\method{} outperforms training‑free tool-calling systems by a wide margin as seen in \cref{tab:main}. On A-OKVQA, we see $19.0\%$ and $21.4\%$ improvements over ViperGPT and Chameleon respectively (see \Cref{tab:main} for full results). The gap on both MMMU and NaturalBench compared to gap between \method{} and single/multi-agent baselines. These tool-calling systems plan their tool execution in a single step based solely on the question, and they lack the ability to reason directly on the image as an input. This is in contrast to our own system that effectively incorporates tools on top of existing VLM systems, allowing for better vision understanding with tools while building on the existing capabilities of VLMs. The performance gain of \method{} over LLM-based tool calling methods illustrates the strengths of our targeted tool-calling approach for a debate among multiple VLM agents.

We also see consistent gains across all three benchmarks over all baselines.
On A-OKVQA, \method{} reaches $68.9\%$, exceeding the best single‑agent baseline (Self‑Consistency with Ovis2) by $4.3\%$ and the best multi‑agent baseline (multi‑model debate with judge) by $3.4\%$.
This trend continues across both MMMU and NaturalBench, where \method{} consistently outperforms both the strongest single and multi-agent baselines. 
The gains are highest on A-OKVQA, which requires advanced reasoning and world knowledge, indicating that our system improves the overall reasoning process of the answering agents. 
These improvements indicate that resolving disagreements with expert tools and using tool‑aligned agreement scores yields better reasoning and perception than prior single‑ or multi‑agent approaches.

\paragraph{\method{} can adapt to diverse domains by adding appropriate expert tools.}
To show how \method{} can adapt to domains where its current pool of experts is unsuited for a task, we test on M3D with an additional medical expert. If \method{} is able to adapt to new experts, we expect the accuracy to increase, as the medical tool provides valuable info to guide a better discussion.
In \cref{tab:m3d}, our method outperforms the baselines by $1.54\%$ even given the disparities in individual model strengths on this task (i.e., Ovis2 is $10.0\%$ weaker than QwenVL on this dataset). While multi-agent debate underperforms compared to the strongest single-agent baseline by $2.72\%$, \method{} can effectively use the medical expert and multiple agent reasoning paths to bridge this gap and improve even further over the next best method (Qwen2.5-VL) by 1.5\%.
This indicates that \method{} can effectively incorporate new experts, allowing it to adapt to new tasks with minimal implementation effort.

\paragraph{\method{} leads to more diverse and fruitful discussion compared to existing multi-agent methods.} To evaluate the diversity and novelty of ideas during discussion, we measure the textual overlap between agents’ initial responses and their responses after one round of discussion, comparing \method{} with the standard multi-agent debate. For \method{}, tool outputs are appended to the initial responses, enabling the metrics to account for all available information during the discussion round. 
As shown in \cref{tab:text_overlap}, \method{} exhibits substantially lower overlap across all metrics (ROUGE-1/2/L \citep{lin-2004-rouge} and Jaccard Index), with ROUGE-L and Jaccard Index decreasing by 0.173 and 0.289 respectively compared to multi-agent debate (MAD) \citep{du2024improving}. This indicates that discussions in \method{} are more diverse, with agents introducing novel reasoning paths rather than repeating prior content. The integration of tool-based resolution enables each agent to refine or challenge other perspectives with visual or textual evidence, resulting in richer discussion. In contrast, multi-agent debate seems to converge in terms of word usage, with agents repeating similar or nearly identical reasoning. These findings show that \method{} fosters more fruitful discussion among agents.

\paragraph{Additional Results.} We conduct additional experiments to study calibration, multi-round performance, efficiency, tool benefits, and prompt sensitivity in \method{}. This additional analysis is included in \cref{app:additional_results}.

\subsection{Analysis and Ablation Study}

\begin{figure}[t]
\begin{center}

\includegraphics[width=0.95\columnwidth]{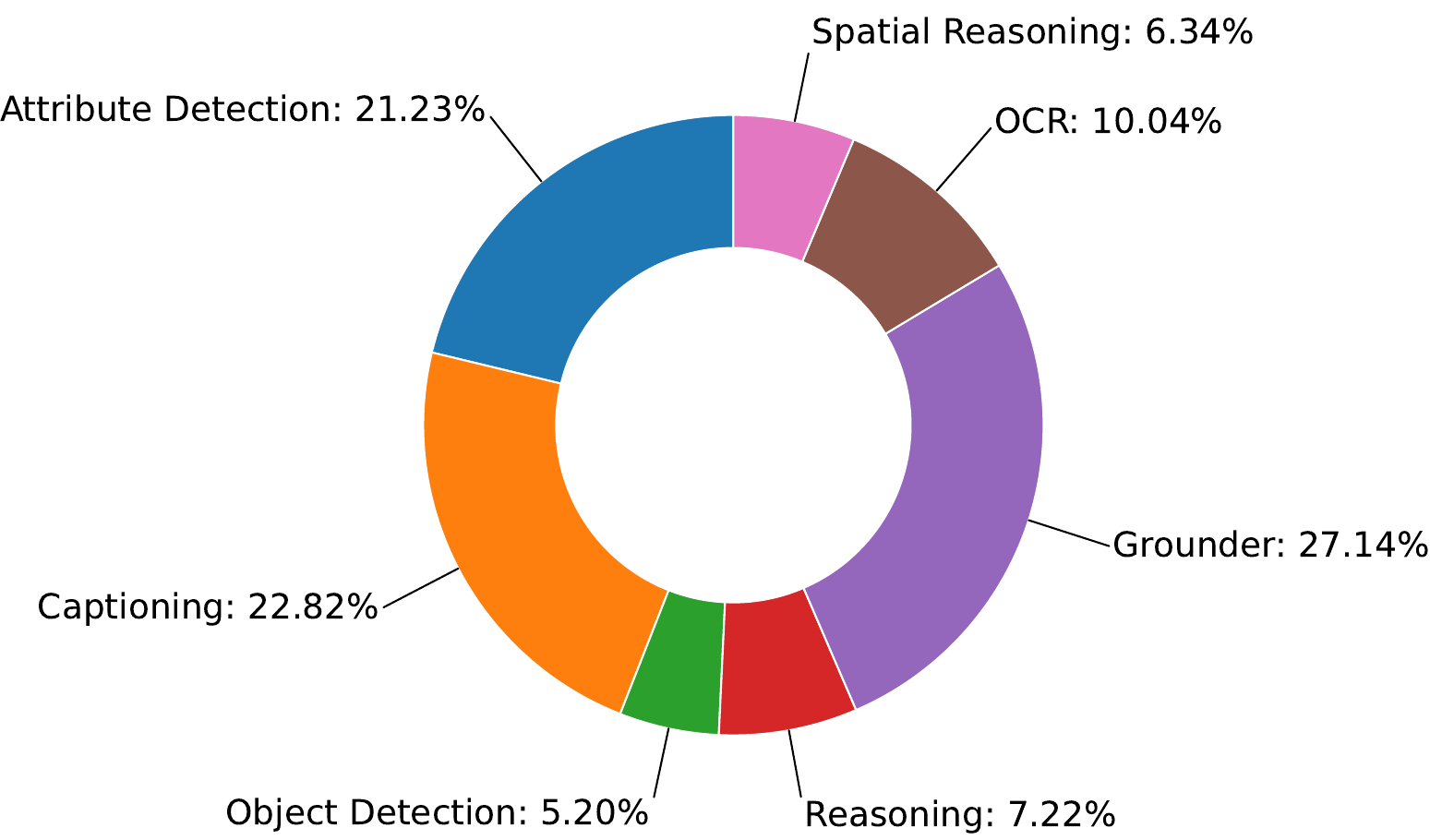}
\end{center}

\caption{Breakdown of total tool calling of \method{} on A-OKVQA.
}
\vspace{-0.5em}
\label{fig3:tool_distribution}
\end{figure}

\paragraph{\method{} calls on a variety of tools.} In \Cref{fig3:tool_distribution}, we present the distribution of tool calls made by \method{} when evaluated on A-OKVQA. The most common tools called were Grounder, Captioning, and Attribute Detection, with each called for over $20\%$ of questions with tool calls. These tools are mostly focused on physical scene understanding and composition, indicating that the VLM agents often differ in their vision understanding capabilities. The less called tools were OCR, Reasoning, Spatial Reasoning, and Object Detection respectively, with each called for $\leq$ 10\% of questions. These are less holistic, more fine-grained vision skills that may not be needed for most questions. Overall, every tool is being invoked and contributing towards more detailed vision understanding.

\paragraph{Multiple different answerers lead to more disagreements.} We compare using multiple models (i.e., QwenVL, MiniCPM-o, and Ovis2) as answers to multiple instances of the same model (i.e., Ovis2 with higher temperature). From \cref{tab:main}, we see that using a single model in \method{} results in a decrease of 4.6\% accuracy on A-OKVQA. Then, we find that using a single model results in less than half the disagreements than \method{} with multiple models, where we count 0.49 and 1.32 disagreements on average respectively. This indicates that the diverse perspectives inherent in a multi-model system yields a richer disagreement signal that allows for better expert tool calling.

\begin{table}[t]
\centering
\resizebox{\columnwidth}{!}{
\begin{tabular}{l r}
\toprule
\textbf{Method} & \textbf{Accuracy} \\
\midrule
\method{} & \textbf{68.9} \\
w/o Diverse Answerers (only Ovis)         & 65.3 \\
w/o Tool Agreement Scores               & 67.8 \\
w/o Expert Tool Outputs$^\dagger$  & 65.7 \\
\bottomrule
\end{tabular}
}
\vspace{-0.5em}
\caption{Ablations of \method{} on A‑OKVQA. $^\dagger$Agreement scores are impossible to calculate without tool outputs, so they are withheld.}
\label{tab:ablations}
\end{table}

\begin{table}[t]
\centering
\resizebox{\columnwidth}{!}{
\begin{tabular}{l r}
\toprule
\textbf{Method} & \textbf{Accuracy} \\
\midrule
\method{} & 68.9 \\
\method{} (VLM as all tools) & 66.0 \\
\method{} (no VLM-based tools) & 66.7 \\
\method{} (no specialized model tools) & 65.6 \\
\bottomrule
\end{tabular}
}
\vspace{-0.5em}
\caption{Comparison between the full \method{} system and its variant using VLMs as tools. VLM-based tools are the captioning, attribute detection, and reasoning tools. Specialized model tools are the grounding, object detection, spatial reasoning, and OCR tools.}
\label{tab:dart_vlm_comparison}
\end{table}

\paragraph{Effect of using a VLM as an expert tool.} 
\label{app:vlm_effects}
While VLMs exhibit strong general capabilities across perception and reasoning tasks, they often underperform compared to specialized models on domain-specific tasks (i.e., grounding, spatial reasoning, etc.). In our original setting, we use a general VLM for three tasks: captioning, attribute detection, and reasoning. We extend this to check whether a single VLM could indeed serve as a substitute for \textit{all} tools. To quantify the trade-off between specialized models and general VLMs, we replace all specialized tools in \method{} with a single general-purpose VLM (InternVL2.5) prompted to handle OCR, spatial reasoning, grounding, and object detection on top of its original tasks. As shown in \cref{tab:dart_vlm_comparison}, this setting results in a 2.9\% decrease in overall accuracy on A-OKVQA. Although VLMs are a convenient solution, their generality may come at the cost of reliability particularly for fine-grained visual understanding tasks.

We also measure the effect of ablating each tool type in this heterogeneous tool set: VLM-based tools and specialized model tools. Removing the VLM-based tools (captioning, attribute detection, reasoning) lowers accuracy from 68.9\% to 66.7\%, suggesting that the VLM tools are still able to provide a complementary signal for a domain-specific task even without designated expert capabilities.
Removing only specialized model tools (grounder, object detection, spatial reasoning, OCR) while keeping VLM-based tools lowers accuracy to 65.3\%, showing the importance of using specialized models in the tool pool. 
Overall, these ablations indicate that the strongest performance occurs when combining specialized model tools with VLM-based tools, with specialized expert tools contributing the larger share of the performance gains.

\paragraph{\method{} component ablations.}
 We assess the impact of key components in our system, including diverse answering agents, expert tools, and tool-aligned agreement scores (results in \cref{tab:ablations}).
 We conduct the ablation studies on A-OKVQA as it is specifically designed to test commonsense reasoning and needs precise visual understanding for questions reliant on world knowledge.
Removing the multiple agent answerers or vision-expert components degrades performance compared to the full \method{} system.  
We evaluate \method{} without diverse answerers and with only the strongest performing model (i.e., Ovis2) on A-OKVQA. Using only the strongest single model (Ovis2) on A-OKVQA (i.e., without diverse answerers) leads to a decline in accuracy from 68.9\% to 65.3\%, highlighting the value of model diversity. 
This indicates that \method{} benefits from having the diverse reasoning and expanded knowledge base of multiple VLMs.  Furthermore, retaining a diverse set of VLM agents without expert confidence yields a 1.1\% performance drop,
showing that the calibrated confidence from the experts improves the discussion beyond regular self-reported confidence. Additionally, we test the discussion framework without vision expert outputs.
This setting is almost equivalent to Debate with Judge (from \Cref{tab:main}) but with only one debate round. Accuracy drops by 3.2\% in this experiment. This result suggests that visual information from experts leads to a higher-quality discussion compared to that of only answerers. In fact, the performance dip is comparable to that of having no diverse answerers, indicating that they are relatively equally important to the overall system. The ablations highlight the importance of having multiple VLMs and vision experts working together in discussion.

\section{Related Work}
\label{sec:related_work}

\paragraph{Tool-Calling Systems.} Past efforts have augmented large language models with external tools to overcome knowledge and reasoning limitations, especially in vision-language tasks \citep{surismenon2023vipergpt, lu2023chameleon, qin2024toolllm, gupta2022visualprogrammingcompositionalvisual, chungen2025hydraagenticreasoningapproach, yang2023mmreact, shen2023hugginggpt, schick2023toolformerlanguagemodelsteach, parisi2022talmtoolaugmentedlanguage}. This tool augmentation has enabled LLMs to tackle tasks such as finding real-time information through web search \citep{lu2023chameleon, nakano2022webgptbrowserassistedquestionansweringhuman}, accessing vision information \citep{lu2023chameleon, surismenon2023vipergpt, wu2023visualchatgpttalkingdrawing, yang2023mmreact, gupta2022visualprogrammingcompositionalvisual}, and acquiring expert domain-specific knowledge \citep{das2024mathsenseitoolaugmentedlargelanguage, yu2025mexageneralmultimodalreasoning}. One direction has used a programmatic approach by leveraging LLMs inherent code generation capabilities to handle tool calls \citep{shen2024pyramidcoderhierarchicalcode, surismenon2023vipergpt} while other work has experimented with directly prompting an LLM to construct a series of tool calls \citep{lu2023chameleon, chungen2025hydraagenticreasoningapproach, yang2023mmreact}. While past work has mostly focused on using a single LLM agent for tool calling, we investigate how using disagreements in multimodal multi-agent debate can lead to more effective tool calling and discussion.
\paragraph{Multi-Agent Systems.}
Another line of research improves reasoning by enlisting multiple LLM agents that collaborate or compete among one another to arrive at a stronger conclusion \citep{chen2024reconcileroundtableconferenceimproves, estornell2025acccollabactorcriticapproachmultiagent, qian2024chatdevcommunicativeagentssoftware, zhao2024electoralapproachdiversifyllmbased, tang2024codeagentautonomouscommunicativeagents, yin2024pearrobustflexibleautomation, zhang2024proagentbuildingproactivecooperative, du2023improvingfactualityreasoninglanguage}. A large body of work has explored designing a debate framework for agents to interact and refine their answers through multiple rounds of discussion \citep{chen2024reconcileroundtableconferenceimproves, estornell2024multillm, du2023improvingfactualityreasoninglanguage}. Others have tried assigning agents specific functions or roles to accomplish different goals or subtasks for a given query \citep{estornell2025acccollabactorcriticapproachmultiagent, hong2024metagpt, wang2024unleashingemergentcognitivesynergy}. Past work has also tested how to control the level of disagreement in a debate \citep{khan2024debatingpersuasivellmsleads, liang2024encouragingdivergentthinkinglarge, chang2025evinceoptimizingmultillmdialogues}.
Our method contributes several novelties along a number of axes. 
First, prior multi-agent debate methods have largely omitted tool use while we inject novel information via disagreement-triggered tool use. 
Moreover, we expand this into a less explored frontier by examining multimodal agents instead of text. 

\section{Conclusion}
\label{conclusion}

In this work, we introduce \method{}, a multi-agent framework that recruits domain-specific vision tools based on disagreements among VLM agents. By identifying points of disagreement in agent discussions and selectively invoking expert tools to resolve these disagreements, \method{} combines the reasoning flexibility of VLMs with the specialized perception capabilities of expert tools. Through this, \method{} sees consistent improvements across four diverse benchmarks over strong baselines on A-OKVQA, MMMU, NaturalBench, and M3D. \method{} represents a promising direction for multimodal reasoning that harnesses the strengths of generalist models and specialized tools through disagreement resolution.

\section*{Acknowledgments}
This work was supported by NSF-CAREER Award 1846185, DARPA ECOLE Program No. HR00112390060, ONR Grant N00014-23-1-2356, ARO Award W911NF2110220, Microsoft Accelerate Foundation Models Research (AFMR) grant program, NSF AI Engage Institute DRL-2112635, National Institutes of Health (NIH) under other transactions 1OT2OD038045-01, and Cisco and Capital One Faculty Awards. The views and conclusions contained in this document are those of the authors and should not be interpreted as representing official policies, either expressed or implied, of the NIH or other sponsors.

\bibliography{custom}

\appendix

\section{Model Configurations}
We use the default generation configurations from HuggingFace for Qwen2.5-VL 7B (Apache-2.0 license), MiniCPM-o 2.6 8B (Apache-2.0 license), and Ovis2 8B (Apache-2.0 license). Qwen2.5-VL uses a temperature of $10^{-6}$ and repetition penalty of 1.05. MiniCPM-o and Ovis2 both use a temperature of 0.7 and repetition penalty of 1.05. For instances of multi-agent debate and \method{} where we would like more answer diversity with a single model, we increase temperature to 0.9 for all agents.

We use the \href{https://huggingface.co/IDEA-Research/grounding-dino-base}{\texttt{grounding-dino-base}} implementation in HuggingFace of GroundingDINO (Apache-2.0 license) with box and text thresholds of 0.3. We use OCR-Qwen (Apache 2.0 License), a Qwen2.5 7B model finetuned for OCR, with default generation configuration of temperature 0.1 and repetition penalty 1.05. We use SpaceLLaVA (Apache-2.0 License) for spatial reasoning with a temperature of 0.2 and repetition penalty of 1.1. InternVL2.5-MPO 8B (MIT License) uses temperature 0.8 for all three expert tool functionalities.

\section{Baseline Details} 
\label{app:baselines}
We compare \method{} to multiple single-agent and multi-agent baselines. The single agent baselines include zero-shot CoT \citep{wei2022chain}, self-refinement \citep{selfrefine}, and self-consistency \citep{wang2023selfconsistency}. In zero-shot COT, we prompt models to first output their reasoning before providing their final answer. With self-refinement, models are asked to provide feedback on their own output and update their answer accordingly. This assesses whether models can improve themselves based on their own feedback. With self-consistency, we sample multiple reasoning paths from a model and take the most consistent answer. This takes advantage of the existence of multiple reasoning paths to arrive at an answer by increasing the number of generated samples per question.

We include comparisons to training-free single-agent tool-calling baselines: ViperGPT \citep{surismenon2023vipergpt} and Chameleon \citep{lu2023chameleon}. ViperGPT utilizes a code generation model to predict executable code that leverages required tools for a given question. In contrast to the flexibility offered by free-form code generation in ViperGPT, Chameleon adopts a more robust compositional approach, utilizing an LLM-based planner to directly invoke various tools in sequence via word matching.
We update both implementations, using Qwen2.5 7B Instruct as the code generation model and incorporating our expert tools into both systems.
For ViperGPT specifically, we also update its VLM module (called on by the code generation model) from BLIP2 \citep{blip2} to InternVL-2.5 MPO for a fairer comparison with our system. 

Finally, we also evaluate against strong multi-agent baselines designed for text-based tasks, including both consensus-based debate and debate with a judge. We follow the ReConcile LLM debate framework outlined by \citet{chen2024reconcileroundtableconferenceimproves} for both consensus and InternVL2.5 as a judge. In the consensus setting, the majority -- or consensus -- answer is automatically taken (a random answer is taken if all agents disagree). In the judge setting, a judge model is given the image along with each answer, justification, and confidence before picking a single final answer.

\begin{table}[h]
    \centering
    \begin{tabular}{clc}
      \toprule
      \textbf{Tool Conf.?} & \textbf{Metric} & \textbf{ECE}$\downarrow$ \\
      \midrule
      \ding{55}  & Self-Reported  & 0.4581 \\
      \midrule
      \ding{51} & Self-Reported  & \textbf{0.3859} \\
      \ding{51} & Tool-Confidence & 0.4901 \\
      \bottomrule
    \end{tabular}
    \caption{Expected calibration error (ECE) of \method{} with and without tool confidence on self‑reported confidence metrics and our tool‑based agreement score metric.}
    \label{tab:conf}
\end{table}
\section{Dataset Details}
\begin{itemize}
    \item \textbf{A-OKVQA} (Apache-2.0 License): We evaluate on the validation split of 1135 samples.
    \item \textbf{MMMU} (Apache-2.0 License): We evaluate on the validation split of 900 samples.
    \item \textbf{NaturalBench} (Apache-2.0 License): We evaluate on the full dataset of 7600 question-image pairs.
\end{itemize}

\section{Additional Results}
\label{app:additional_results}

\paragraph{\method{} provides a stronger confidence metric based on vision tools.} \label{app:conf} We test the effectiveness of calculating confidence based on agreement to vision tools. For this, we run \method{} under two settings: 1) using only self-reported confidence, and 2) combining self-reported confidence with tool-based agreement scores. Both settings still involve disagreement resolution with tools. We then measure the effectiveness using expected calibration error (ECE) \citep{ece} of self-reported confidence only, tool-based confidence, and self-reported confidence given tool-based confidence during discussion. 
Recall that both self-reported confidence and tool-based confidence are used in \method{} during discussion. 

We observe higher ECE when relying solely on self‑reported confidence (i.e., without tool‑based agreement scores) than when including those scores: $0.4581$ vs. $0.3859$ respectively (see \cref{tab:conf}).
Note however that the raw tool-based confidence calculated in \method{} has ECE of $0.4901$, which indicates weaker calibration than self-reported confidence on its own.

\begin{table}[h]
    \centering
    \resizebox{\columnwidth}{!}{
    \begin{tabular}{ccccc}
      \toprule
      \textbf{Round} & \textbf{QwenVL} & \textbf{MiniCPM} & \textbf{Ovis} & \textbf{w/ Aggregation} \\
      \midrule
      Initial & 61.3   & 55.7   & 61.0   & 59.4   \\
      1       & 68.6   & 66.8   & $\mathbf{69.2}$ & $\mathbf{68.9}$ \\
      2       & $\mathbf{68.7}$ & 67.2   & 68.7   & 68.4   \\
      3       & 68.4   & $\mathbf{67.4}$ & 68.7   & 68.5   \\
      \bottomrule
    \end{tabular}
    }
    \caption{\method{} performance over three rounds of disagreement resolution and expert confidence on A-OKVQA.
    }
    \label{tab:rounds}
\end{table}
\paragraph{Multiple rounds of debate in multimodal multi-agent systems is ineffective.} \label{app:multiround}

\begin{figure}[t]
\begin{center}

\includegraphics[width=0.9\columnwidth]{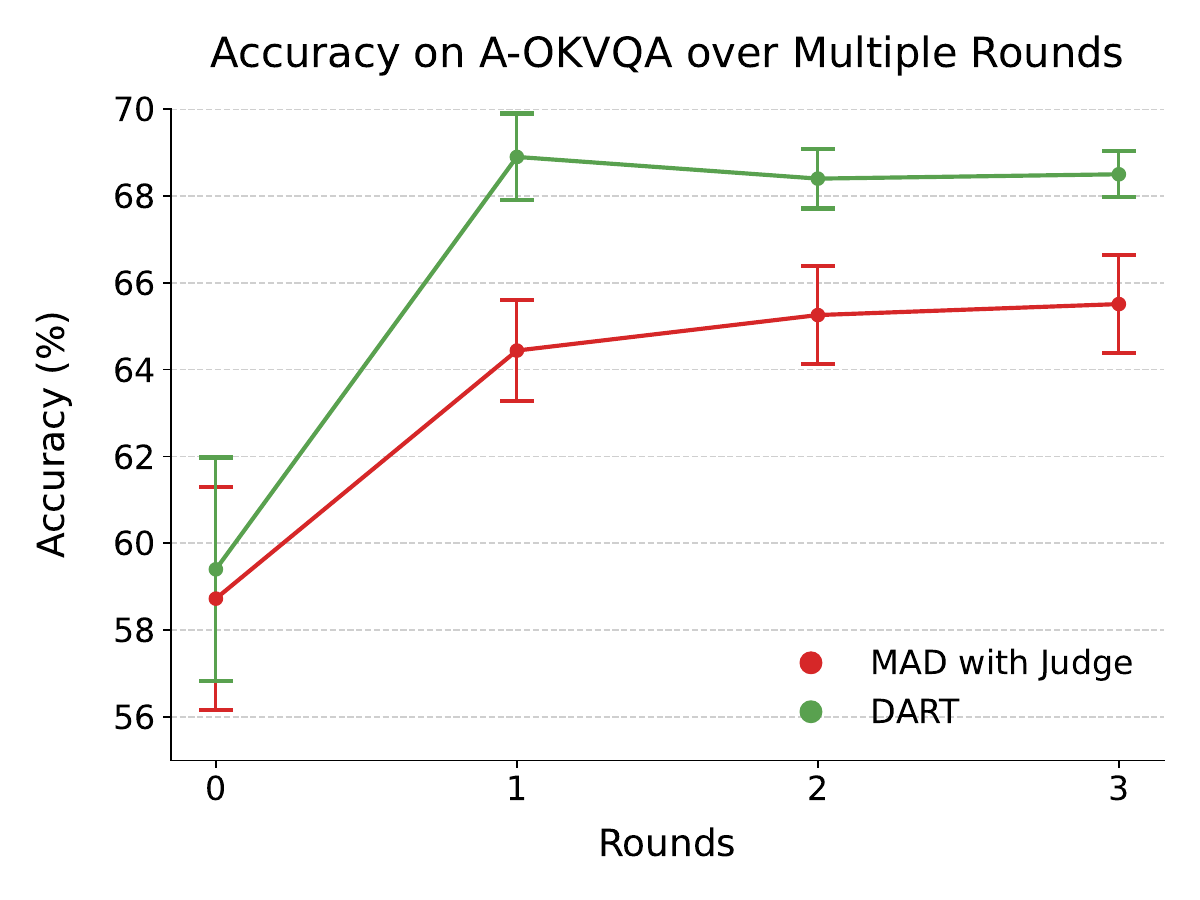}
\end{center}

\caption{Performance of \method{} and multi-agent debate over three rounds. The error bars indicate the standard deviation among the individual answering agents.}
\label{fig:rounds}
\end{figure}
Encouraged by the success of multiple debate rounds in text-based multi-agent environments, we apply our framework across successive debate rounds to see if the same pattern emerges. We compare the improvement of \method{} to multi-agent debate with judge over three rounds  in \cref{fig:rounds}. \method{} after one round of expert tool intervention outperforms the round three performance of multi-agent debate with judge. Both methods have the largest improvement in round after the initial answer generation. The improvement is likely due to initial exposure to other agent outputs and, in the case of \method{}, tool outputs. However, after the first round of discussion both stagnate in subsequent rounds (with <2\% accuracy gains between rounds 1 and 3). For multi-agent debate, the agents are not able to effectively refine their reasoning and answer after round one as there is no new information being introduced. For \method{}, we observe that the most major disagreements are resolved in round 1. Future rounds are prone to resolving more minor disagreements that can distract from the question. We find that it is insufficient to give any piece of new information in a multimodal multi-agent system. In fact, the new expert tool information must be sufficiently good and relevant to advance discussion.

\begin{table}[h]
    \centering
    \begin{tabular}{lcc}
      \toprule
      \textbf{Model} & \textbf{All Choices} & \textbf{Accuracy} \\
      \midrule
      InternVL2.5 & \ding{55}    & $68.3$ \\
      Qwen2.5-VL  & \ding{55}      & \underline{$68.7$} \\
      MiniCPM-o 2.6 & \ding{55}  & $67.8$ \\
      \midrule
      Ovis2          & \ding{51} & $68.5$ \\
      Ovis2          & \ding{55} & $\mathbf{68.9}$ \\
      \bottomrule
    \end{tabular}
    \caption{Aggregator Ablation on A-OKVQA.
    All choices indicates whether answers from initial generation are used. } 
    \label{tab:judge_results}
\end{table}

\paragraph{Robustness of \method{} aggregator to model choice.} 
We test four different VLMs as aggregator models: Qwen2.5-VL, MiniCPM-o, Ovis2, and InternVL-2.5 MPO. 
The results are shown in \cref{tab:judge_results}. 
Between the strongest and weakest aggregator (Ovis2 and MiniCPM-o, respectively), there is a difference in accuracy of $1.1\%$. In fact, among the top 3 models, accuracies are within $0.6\%$ of each other. This indicates \method{} has relatively low sensitivity to aggregator choice and will generalize well to new models as aggregators. 
Additionally, we test how the aggregator model responds when answer choices from the initial generation stage are provided in addition to the discussion stage answers. 
We find that including these additional answers slightly lowers performance by 0.4\%. This suggests that having a more concise answer set after using tools (i.e., post discussion answers) makes aggregation easier compared to lower quality answers from initial generation.

We also find that the tool-based agreement scores are beneficial in improving calibration during aggregation (see \cref{tab:conf} in the Appendix.). When aggregating with agreement scores, the final self-reported confidence values in \method{} are better calibrated and obtain a relative decrease of 15.8\% in expected calibration error (ECE) compared to self-reported confidence values without tool-based agreement scores. More discussion on the calibration calculations is in \cref{app:conf}.

\begin{table}[h]
\centering
\resizebox{\columnwidth}{!}{
\begin{tabular}{l r}
\toprule
\textbf{Method} & \textbf{Accuracy} \\
\midrule
\method{} & \textbf{68.9} \\
w/o Aggregation (i.e., Majority Vote) & 67.9\\
w/o Discussion & 62.5 \\
\bottomrule
\end{tabular}
}
\caption{Ablations of discussion and aggregation steps of \method{} on A‑OKVQA.}
\label{tab:extra_ablations}
\end{table}

\paragraph{Additional Method Ablations.} We observe that removing aggregation and instead using a simple majority vote leads to reduced accuracy by 1.0\% in \cref{tab:extra_ablations}, confirming that the aggregation step is more effective than naive majority vote. In fact, majority vote underperforms aggregation in all but one setting tested in \cref{tab:judge_results}, demonstrating that proper processing of agent reasoning and tool outputs creates better final answers. Finally, removing the discussion phase causes the largest performance drop by 6.4\%, emphasizing its role in refining the preliminary answers. In fact, the quality of answers after the initial reasoning phase is substantially higher (up to 11.1\% higher for the weakest agent as reported in \cref{tab:rounds}). This indicates the discussion step allows for stronger reasoning and responses, yielding to a more accurate downstream system output. The ablations emphasize the importance of incorporating multiple VLMs, expert tool information, and discussion steps to the system.

\begin{table}[h]
    \centering
    \resizebox{\columnwidth}{!}{
    \begin{tabular}{lr}
      \toprule
      \textbf{Model} & \textbf{Avg. Tokens Generated} \\
      \midrule
      Ovis2    & $74.3$ \\
      MiniCPM-o    & $36.7$ \\
      QwenVL        & $80.5$ \\
      MAD with Consensus & $1031.8$ \\
      MAD with Judge  & $1196.8$ \\
      \midrule
      \method{}  & $928.5$ \\
      \bottomrule
    \end{tabular}
    }
    \caption{We calculate the average tokens generated of each method on A-OKVQA. This represents the relative cost of each method.}
    \label{tab:tokens_generated}
\end{table}

\begin{table}[h!]
\centering
\resizebox{\columnwidth}{!}{
\begin{tabular}{lccc}
\toprule
\textbf{Method} & \textbf{Total Runtime (hrs)} & \textbf{TFLOPs} & \textbf{Latency (s)} \\
\midrule
Ovis2 & 0.13 & 21.17 & 0.38 \\
MiniCPM-o & 0.09 & 18.92 & 0.26 \\
QwenVL (7B) & 0.05 & 19.04 & 0.16 \\
QwenVL (32B) & 0.20 & 88.01 & 0.64 \\
MAD w/ Consensus & 7.97 & 215.66 & 21.84 \\
MAD w/ Judge & 8.88 & 236.83 & 27.92 \\
\method{} & 10.52 & 158.15 & 32.75 \\
\bottomrule
\end{tabular}
}
\caption{Comparison of system efficiency across different methods on A-OKVQA.}
\label{tab:dart_efficiency}
\end{table}

\paragraph{\method{} provides a balance between performance and computation compared to both single and multi-agent baselines.}\label{app:cost} We investigate the cost of using \method{} compared to single and multi-agent baselines. We first count the average number of tokens generated per method on A-OKVQA and present the results in \cref{tab:tokens_generated}. As expected, single agent methods with CoT are the cheapest, with QwenVL (the most token-heavy single agent baseline) generating 80.5 tokens compared to \method{}'s 928.5. However, compared to multi-agent systems, \method{} needs 103.3 fewer tokens on average than a multi-agent debate system.

We further assess computational efficiency in \cref{tab:dart_efficiency}, reporting total runtime, floating-point operations (TFLOPs), and latency per query. While \method{} incurs higher runtime and latency than single-model systems such as QwenVL, its cost remains competitive with other multi-agent methods. In fact, in terms of FLOPs, \method{} requires fewer FLOPs than the multi-agent baselines (although still behind in total runtime and latency). The observed difference in runtime and latency between multi-agent debate and \method{} is likely due to inefficient tool-calling implementation relative to discussion/answering implementations.

Additionally, we test a larger VLM (Qwen2.5-VL 32B) to compare efficiency and performance when scaling versus using multiple agents. While Qwen2.5-VL 32B is more cost-effective, its performance gains are modest (scoring 62.7\% on A-OKVQA). Although pricier, multi-agent approaches like DART may offer a viable path to improved results when scaling yields diminishing returns. Overall, \method{} achieves stronger performance from the incorporation of tools at the cost of reduced efficiency compared to multi-agent methods and single-agent methods.

\begin{table}[h!]
\centering
\resizebox{0.9\columnwidth}{!}{
\begin{tabular}{l r}
\toprule
\textbf{Tool Witheld} & \textbf{Accuracy} \\
\midrule
\method{} & 68.9 \\
w/o Grounder & 66.5 \\
w/o Captioning & 67.5 \\
w/o Attribute Detection & 68.6 \\
w/o OCR & 68.1 \\
w/o Reasoning & 68.4 \\
w/o Spatial Reasoning & 68.0 \\
w/o Object Detection & 68.9 \\
\bottomrule
\end{tabular}
}
\caption{Ablation for tools of DART on A-OKVQA. We measure performance impact on the entire system when we remove a single tool from the tool set. Tools are ordered in terms of most to least frequent as reported in \cref{fig3:tool_distribution}.}
\label{tab:tool_ablations}
\end{table}

\paragraph{Each tool complements \method{}'s performance.} We seek to test whether \method{} is able to effectively use the tools that it has been equipped with. From \cref{tab:tool_ablations} we can infer the relative importance of each tool to DART’s overall performance on A-OKVQA. In fact, we observe that the performance drops roughly correlate with the tool usage frequency recorded in \cref{fig3:tool_distribution}. More specifically, the \textit{grounder} and \textit{captioning} tools -- each contributing to over 20\% of tool invocations on A-OKVQA -- are the most critical, as their removal leads to a degradation of more than 2\% in accuracy. The remaining tools (\textit{OCR}, \textit{spatial reasoning}, \textit{attribute detection}, and \textit{reasoning}) yield smaller decreases, typically within 1\%, suggesting that \method{} remains relatively robust to their absence. Interestingly, the omission of object detection does not noticeably affect accuracy, implying that visual grounding and caption-based context already provide sufficient perceptual cues for most questions. Overall, these results highlight that while certain tools (particularly grounding and captioning) play a dominant role, the complementary interaction among all tools contributes to the system’s strong performance, demonstrating that \method{} effectively leverages its tool reasoning capabilities.

\paragraph{\method{} is robust to prompt choice.} We test \method{}'s performance under an altered set of prompts to test the system's sensitivity to prompt choice. Each stage's prompt is passed into GPT-5 \citep{openai2025gpt5} with the prompt ``I am creating a prompt for a system that uses tools and agents. Rephrase the following prompt:'' while we manually adjust the newly generated prompt to ensure the output format and handling of string variables is consistent. Testing this new set of prompts on A-OKVQA, we achieve accuracy of 68.7\%, a 0.2\% decrease from our previous baseline of 68.9\%. While there is a dip in performance, the change is relatively small, and our method still outperforms the other baselines.

\section{Prompts}
We provide the prompts used in our methods for reproducibility. The details on how these prompts are used are provided in \cref{sec:method}.

\begin{figure*}[h]
\begin{tcolorbox}[
    fontupper=\scriptsize,
    fontlower=\scriptsize,
    colback=white,
    colframe=gray,
    title=\textbf{Initial Answer Generation Prompt}, 
    fonttitle=\bfseries\small, 
    arc=4mm, 
]
<question>\\

Answer the question with only one word or phrase, then provide step-by-step reasoning for the answer. Finally, provide a confidence score from 0-1 for your answer (0 meaning not confident at all, 1 meaning complete confidence)\\

You must output in the following format:

Answer: [answer]

Reasoning: [reasoning]

Confidence: [confidence]
\end{tcolorbox}
\vspace{-1.5em}
\label{fig:initial}
\end{figure*}

\begin{figure*}[t]
\begin{tcolorbox}[
    fontupper=\scriptsize,
    fontlower=\scriptsize,
    colback=white, 
    colframe=gray, 
    title=\textbf{Disagreement Resolution Prompt}, 
    fonttitle=\bfseries\small, 
    arc=4mm, 
]

Here was the initial prompt: <full initial answer generation prompt>\\

Carefully review the following solutions from other agents for the provided question. Now, analyze what disagreements are occurring between the different agents.\\

<grouped solution>

You have the ability to call on different experts, each with their own specialized capabilities. Based on the disagreements you observed, pick out the set of experts (could be just one) that would be best equipped to solve all the disagreements.\\

Here are all the experts, their inputs, and their capabilities/usage:\\

"spatial" (input: list. objects that have confused spatial relations) - Has perfect understanding of spatial relations between objects. Use this when agents are unsure about the placement of items in a scene.
"ocr" (input: none)- Can correctly read all text in an image. Use this when agents have differing views on what the text is in an image.
"grounder" (input: list. objects you are trying to find) - Will find any object if it is an image, otherwise it will return nothing. Use this when agents are not agreeing on what's present in an image.
"detector" (input: none) - Will provide a list of objects in the image, their counts, and their bounding boxes. Only use this when agents are differing in their counts of objects in an image.
"captioning" (input: list. objects you want captions for) - Can give a detailed description of what's going in the image relevant to the question. Use this when agents might need a better idea of the general scene or descriptions of specific objects.
"attribute" (input: list. objects you want attributes for) - Will give information on different features of objects in the image, including color, properties, catgories, and more. Use this when agents are confused about the features of relevant objects and need many surface level features.
"reasoning" (input: list. objects you want reasoning for) - Has better world knowledge and advanced reasoning capabilities about what might be going on in an image. Use this when agents are confused or conflicting in their inferences about the scene. This is essentially a meta-reasoning agent that intervenes when models have different conclusions based on the same assumptions.\\

Output the expert(s) you need to resolve the disagreements to answer the original question: {}. This should be a JSON with this format like this. You can call more or less experts than this as needed:\\

\{
\\\hspace*{1em}"experts": ["grounder", "attribute", "ocr"],
\\\hspace*{1em}"inputs": \{
\\\hspace*{2em}"grounder": \{
\\\hspace*{3em}"disagreement": "Agent 1 mentioned that there is a cat, but Agent 2 said there is no cat and instead said it is a dog.",
\\\hspace*{3em}"justification": "The grounder will help resolve the disagreement about the presence of a cat or dog in the image.",
\\\hspace*{3em}"arguments": ["cat", "dog"]
\\\hspace*{2em}\},
\\\hspace*{2em}"attribute": \{
\\\hspace*{3em}"disagreement": "Agent 1 said the flower is red, Agent 2 said it is orange, and Agent 3 did not specifically mention anything about the flower. There also was confusion about the details of the car.",
\\\hspace*{3em}"justification": "The attribute expert will help resolve the disagreement about the color of the flower and provide details about the car.",
\\\hspace*{3em}"arguments": ["flower", "car"]
\\\hspace*{2em}\},
\\\hspace*{2em}"ocr": \{
\\\hspace*{3em}"disagreement": "Agents conflict on the text they see in the image.",
\\\hspace*{3em}"justification": "The OCR expert will see the text in the image and resolve the disagreement.",
\\\hspace*{3em}"arguments": []
\\\hspace*{2em}\}
\\\hspace*{1em}\}
\\\}\\

Now give the expert output in the given format for the previous agent solutions and the question provided above. Do not be redudant on disagreements unless the expert is adding new information that better resolves the disagreement. Do not add an expert unless absolutely necessary.\\

Reminder, the question is: <question>

\end{tcolorbox}
\vspace{-1.5em}
\label{fig:disagreement}

\end{figure*}

\begin{figure*}[t]
\begin{tcolorbox}[
    fontupper=\scriptsize,
    fontlower=\scriptsize,
    colback=white, 
    colframe=gray, 
    title=\textbf{Tool Agreement Prompt}, 
    fonttitle=\bfseries\small, 
    arc=4mm, 
]
You are an aligning agent. That means that you determine whether or not two different outputs are misaligned (0) or aligned (1). An expert output has provided a response to a disagreement between multiple different agents. Your task is to determine whether or not the expert output is aligned with a single agent's output. Reason about the alignment of the agent output with the expert output. Then, output a 0 if the agent output is not aligned with the expert's output, and a 1 if it is aligned.\\

Output Format:\\
\{\\
\hspace*{1em}"reasoning": "Your reasoning for the alignment of the agent output with the expert output.",\\
\hspace*{1em}"alignment": "0 if the agent output is not aligned with the expert's output, and 1 if it is aligned."\\
\}\\

Disagreement: <disagreement>\\
Expert Output: <expert\_output>\\
Agent Output: <agent\_output>\\
\end{tcolorbox}
\vspace{-1.5em}
\label{fig:agreement}

\end{figure*}

\begin{figure*}[t]
\begin{tcolorbox}[
    fontupper=\scriptsize,
    fontlower=\scriptsize,
    colback=white, 
    colframe=gray, 
    title=\textbf{Discussion Prompt}, 
    fonttitle=\bfseries\small, 
    arc=4mm, 
]
Carefully review the following solutions from other agents as additional information, and provide your own answer and step-by-step reasoning to the question. Then, give your confidence in your new answer.\\
                
Clearly state which point of view you agree or disagree with and why.\\

<grouped\_solutions>\\

Here is information about the image from expert(s). You should take any information presented in this as factual unless you truly notice something wrong with the expert info (and make sure to state this).\\

<tool\_outputs>\\

Output your response in the following format:\\

Reasoning: [reasoning for your answer to the original question based on the other agents responses]\\
Answer: [new answer]\\
Confidence: [confidence]

Some example outputs:\\

Reasoning: I agree with Agent 1's reasoning that the answer is A because it aligns with the medical expert's analysis of the CT scan. The medical expert confirmed that the condition described in the question is consistent with the findings in the image.\\
Answer: Option A\\
Confidence: 0.65\\

Reasoning: I disagree with Agent 2's reasoning that the answer is B. The medical expert's analysis suggests that the condition is more likely to be A, as it matches the symptoms described in the question and the findings in the image.\\
Answer: Option A\\
Confidence: 0.95\\

Reasoning: I agree with Agent 1 and 3's reasoning that the answer is C. The medical expert's analysis supports this conclusion, as it aligns with the findings in the image and the symptoms described in the question.\\
Answer: Option C\\
Confidence: 0.30\\

You must follow the provided format no matter what. This rule is unbreakable.
\end{tcolorbox}
\vspace{-1.5em}
\label{fig:discussion}

\end{figure*}

\begin{figure*}[t]
\begin{tcolorbox}[
    fontupper=\scriptsize,
    fontlower=\scriptsize,
    colback=white, 
    colframe=gray, 
    title=\textbf{Aggregator Prompt}, 
    fonttitle=\bfseries\small,
    arc=4mm, 
]
You are an aggregator model that will be given an image, question, and set of answer choices. Your tasks is to select the best final answer for the question. You will also be given various sources of information to help inform your decision. Each answer was generated by a different agent, and these agent will provide their reasoning for why they gave their answer. You will also be given tool outputs from expert models that directly relate to the question and were used to resolve disagreement among the answering agents. Feel free to defer to these expert tool outputs if the answers contradict the info from the tools.\\

The question is: <question>\\

Here are the different agent answers:\\
<grouped\_output>\\

Here are the tool outputs:\\
<tool\_outputs>\\

Now based on the provided information, provide your step-by-step reasoning for selecting the best, most correct answer to the question provided. Then, give your confidence in your selected answer.\\

Output in the following format:\\

Reasoning: [reasoning]\\
Answer: [answer]\\
Confidence: [confidence]\\
\end{tcolorbox}
\vspace{-1.5em}
\label{fig:aggregator}
 
\end{figure*}

\section{Qualitative Examples}
\begin{figure*}[h!]
\begin{center}
\includegraphics[width=0.95\textwidth]{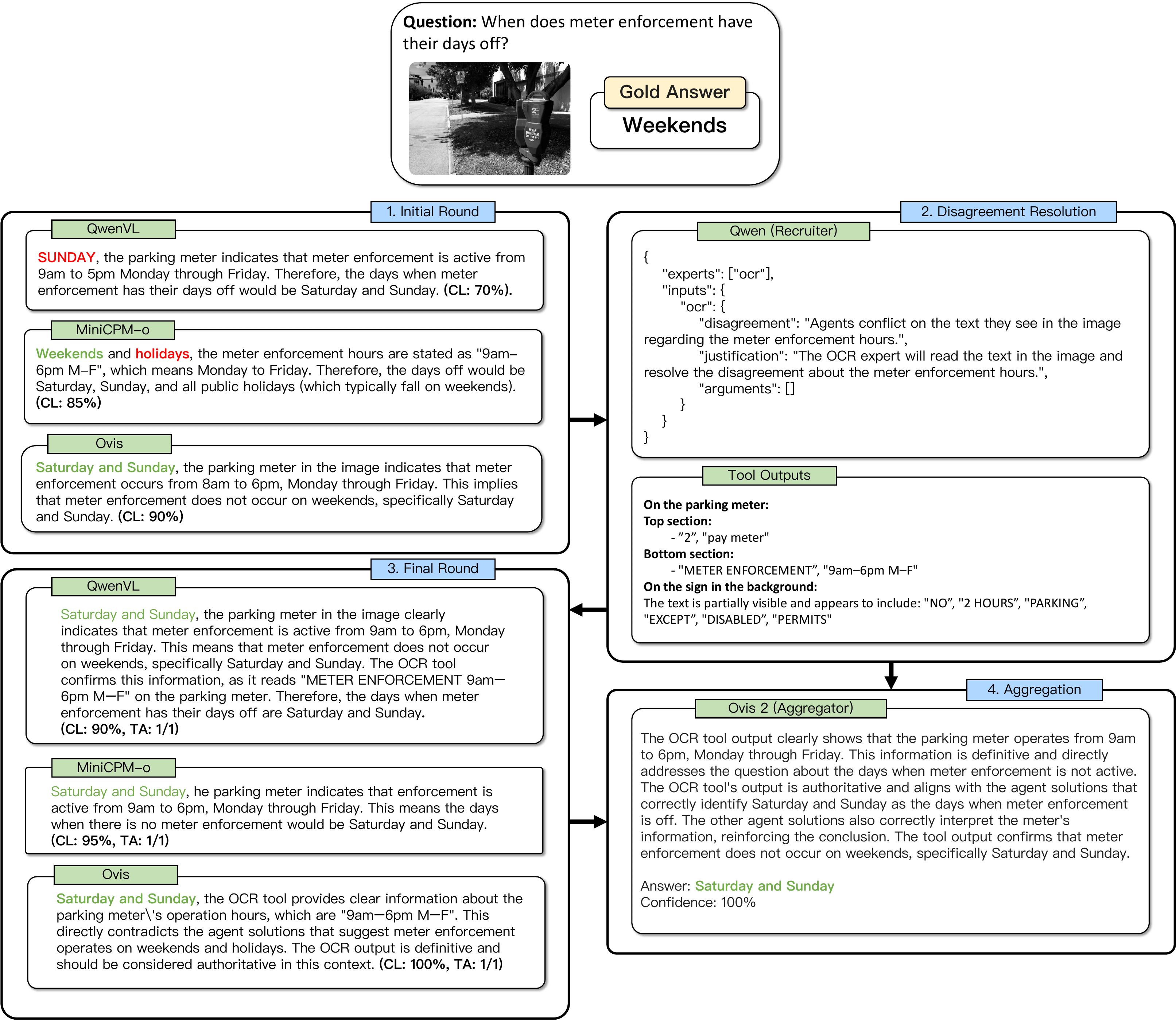}
\end{center}
\caption{Qualitative Example for \method{}. We have input question ``When does meter enforcement have their days off?'' with gold answer \textbf{Weekends}. Ovis is the only model to get it correct, with MiniCPM-o jumping to an improper conclusion and QwenVL being objectively incorrect. The recruiter identifies this disagreement on what text is said on the meter and calls on the OCR tool to resolve the disagreement. The OCR tools correctly identifies the text as M-F 9am-6pm. As a result, the models and aggregators are able to get to the correct answer in subsequent steps.}

\label{fig:qual_example2}
\end{figure*}

\begin{figure*}[h!]
\begin{center}
\includegraphics[width=0.95\textwidth]{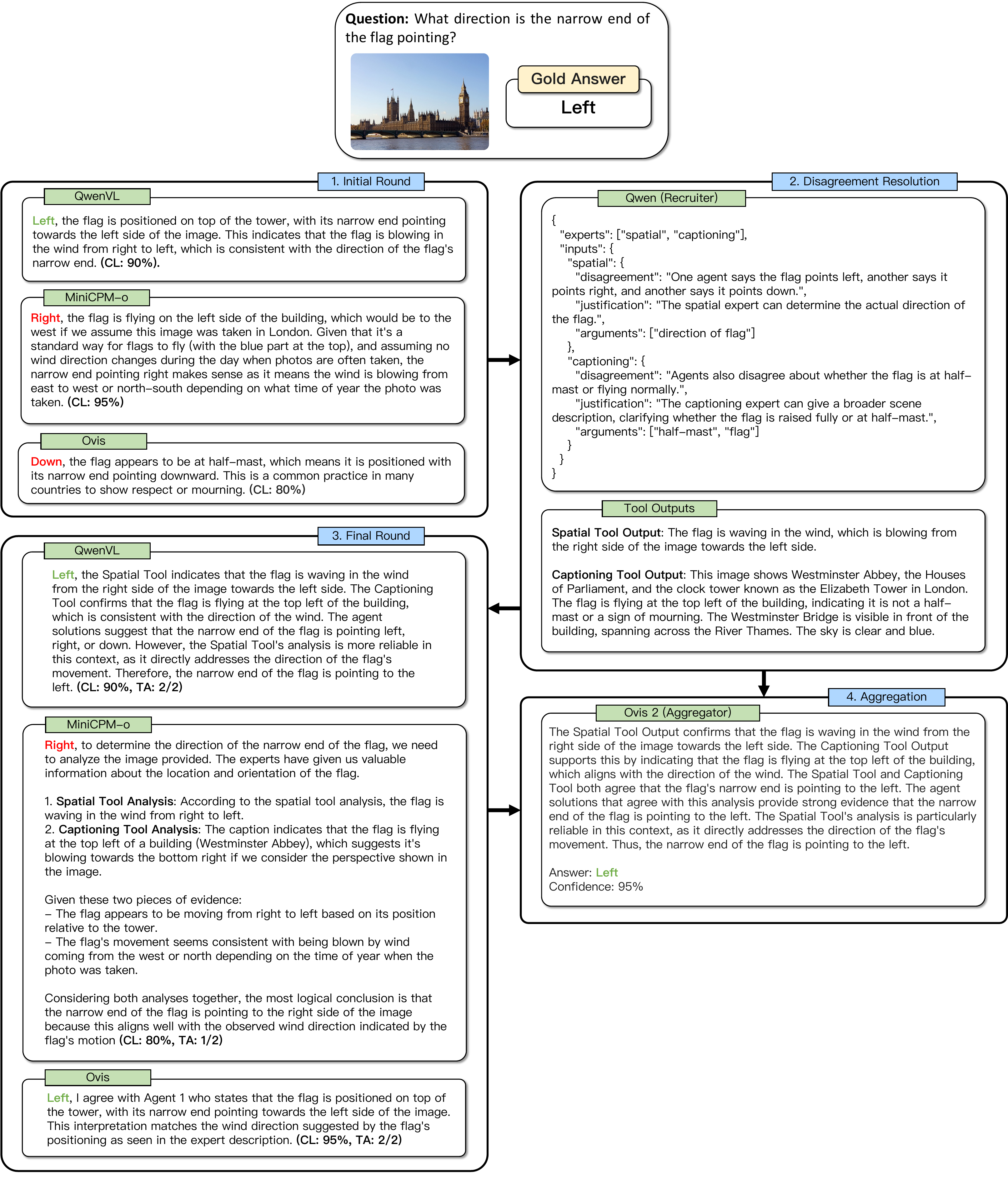}
\end{center}
\caption{Qualitative Example for \method{}. We have input question ``What direction is the narrow end of the flag pointing?'' with gold answer \textbf{Left}. The models all initially disagree, each pointing out different directions (down, left, right). The recruiter identifies this disagreement and calls on the spatial and captioning tools to resolve the disagreement. With the intervention of these tools specifying the direction of the flag, the models refine their answers to get closer to the gold answer, and the aggregator successfully picks the correct answer.}

\label{fig:qual_example1}
\end{figure*}

We provide two qualitative examples showcasing the full \method{} pipeline in \cref{fig:qual_example1} and \cref{fig:qual_example2}.

\end{document}